\documentclass[12pt]{article}

\usepackage[left=1.25in, right=1.25in, top=1in, bottom=1in]{geometry}
\usepackage{times}
\usepackage[utf8]{inputenc}
\usepackage[T1]{fontenc}
\usepackage{amsmath, amssymb, amsfonts}
\usepackage{graphicx}
\usepackage{booktabs}
\usepackage{hyperref}
\usepackage{multirow}
\usepackage{caption}
\usepackage{subcaption}
\usepackage{fancyhdr}
\usepackage{titlesec}
\usepackage{microtype}
\usepackage{enumitem}
\usepackage{needspace}
\usepackage{float}

\titleformat{\section}
  {\bfseries\fontsize{16}{19}\selectfont}{\thesection}{1em}{}
\titlespacing*{\section}{0pt}{2\baselineskip}{\baselineskip}

\titleformat{\subsection}
  {\bfseries\fontsize{14}{17}\selectfont}{\thesubsection}{1em}{}
\titlespacing*{\subsection}{0pt}{2\baselineskip}{\baselineskip}

\titleformat{\subsubsection}
  {\bfseries\fontsize{12}{14}\selectfont}{\thesubsubsection}{1em}{}
\titlespacing*{\subsubsection}{0pt}{2\baselineskip}{\baselineskip}

\usepackage{parskip}
\begin{document}


\vspace*{-0.5in}

\hfill
\raisebox{0.15in}{\fontsize{10}{12}\selectfont MSOE Artificial Intelligence Research Paper}\hspace{0.5em}%
\includegraphics[height=0.5in]{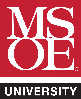}\hspace{0.3em}%
\includegraphics[height=0.5in]{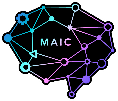}\hspace{-0.5in}

\vfill

\begin{center}
{\fontsize{18}{22}\selectfont SkyNet: Belief-Aware Planning in Partially Observable Stochastic Games}

\vspace{1.5\baselineskip}

{\fontsize{12}{15}\selectfont
Adam Haile
}

\vspace{\baselineskip}

{\fontsize{12}{15}\selectfont
Dwight and Dian Diercks School of Advanced Computing \\
Milwaukee School of Engineering \\
1025 N Broadway St, Milwaukee, WI 53202 \\
\texttt{hailea@msoe.edu}
}
\end{center}

\vfill

\newpage
\pagenumbering{arabic}
\setcounter{page}{1}

\pagestyle{fancy}
\fancyhf{}
\fancyfoot[C]{\thepage}
\renewcommand{\headrulewidth}{0pt}

\begin{center}
{\fontsize{16}{19}\selectfont \textbf{Abstract}}
\end{center}

\vspace{0.5\baselineskip}

In 2019, Google DeepMind released MuZero, a model-based reinforcement learning method that achieves strong results in perfect-information games by combining learned dynamics models with Monte Carlo Tree Search (MCTS). However, comparatively little work has extended MuZero to partially observable, stochastic, multi-player environments, where agents must act under uncertainty about hidden state. Such settings arise not only in card games but in domains such as autonomous negotiation, financial trading, and multi-agent robotics. In the absence of explicit belief modeling, MuZero's latent encoding has no dedicated mechanism for representing uncertainty over unobserved variables.

To address this, we introduce SkyNet (Belief-Aware MuZero), which adds ego-conditioned auxiliary heads for winner prediction and rank estimation to the standard MuZero architecture. These objectives encourage the latent state to retain information predictive of outcomes under partial observability, without requiring explicit belief-state tracking or changes to the search algorithm.

We evaluate SkyNet on Skyjo, a partially observable, non-zero-sum, stochastic card game, using a decision-granularity environment, transformer-based encoding, and a curriculum of heuristic opponents with self-play. In 1000-game head-to-head evaluations at matched checkpoints, SkyNet achieves a 75.3\% peak win rate against the baseline (+194 Elo, $p < 10^{-50}$). SkyNet also outperforms the baseline against heuristic opponents (0.720 vs.\ 0.466 win rate). Critically, the belief-aware model initially underperforms the baseline but decisively surpasses it once training throughput is sufficient, suggesting that belief-aware auxiliary supervision improves learned representations under partial observability, but only given adequate data flow.

\newpage

\section{Introduction}

Game-playing AI has long served as a benchmark for progress in machine learning. From Deep Blue to AlphaZero [2], AI systems have achieved superhuman performance in deterministic, perfect-information two-player games. MuZero [1] extended this line of work by learning a dynamics model entirely from experience, obviating the necessity for knowledge of game rules during planning. These advances were powered by Monte Carlo Tree Search (MCTS) [3], which uses learned policy and value estimates to guide search through a game tree.

However, many games and decision problems are \textit{partially observable}, \textit{stochastic}, and \textit{multi-player}. These properties fundamentally challenge classical MuZero's assumptions. In such settings, an agent cannot observe the full game state, must reason about hidden information, and faces chance events whose outcomes are unknown until they occur. Card games represent an important class of such imperfect-information domains, as demonstrated by landmark results in poker [20, 21, 22].

Skyjo is a multi-player card game (2--8 players) in which each player manages a $3 \times 4$ grid of cards, many of which begin face-down. Players take turns drawing from a shared deck or discard pile, making decisions about which cards to reveal, keep, or replace. The game combines risk management, information gathering, and strategic timing: a player who reveals all their cards triggers the end of a round, but faces a score-doubling penalty if they do not hold the lowest score. Play continues across rounds until a cumulative score threshold is reached.

This combination of hidden information, stochastic draws, multi-player dynamics, and non-zero-sum scoring makes Skyjo a compelling testbed for advancing model-based reinforcement learning beyond the perfect-information regime. The partially observable stochastic domain (POMDP) framework [4] provides the theoretical foundation for reasoning in such settings, where optimal behavior requires maintaining beliefs over hidden state.

This paper makes the following contributions:

\begin{enumerate}[nosep]
    \item A complete, deterministic, seedable Skyjo environment with a decision-granularity action decomposition suitable for MuZero-style training.
    \item Belief-Aware MuZero, an extension of MuZero that adds ego-conditioned winner and rank prediction heads to improve representation learning under partial observability.
    \item An empirical comparison between classical MuZero and Belief-Aware MuZero on Skyjo, demonstrating statistically significant superiority of the belief-aware variant in head-to-head play.
    \item An analysis of training dynamics, stability, and the role of data throughput in realizing the benefits of belief-aware auxiliary supervision.
\end{enumerate}

\section{Related Work}

\subsection{MuZero and Model-Based Reinforcement Learning}

MuZero [1] learns a latent dynamics model purely from interaction data and achieved strong results in Go, Chess, Shogi, and Atari under \textit{perfect information} and deterministic or near-deterministic transitions. The architecture comprises a representation network, a dynamics network, and a prediction network; MCTS [3] uses these to plan in latent space. Grimm et al.\ [14] formalized that models need only be value-equivalent rather than observation-reconstructive, which is relevant in card games where full hidden-state prediction is impossible. MuZero's original formulation does not address partial observability. Extensions such as Stochastic MuZero [7] (chance events and card draws), Gumbel MuZero [8] (limited simulation budgets), and MuZero Reanalyse [9] (target staleness) address other limitations but not hidden information. EfficientZero [10] and self-predictive representation learning [15] show that auxiliary representation-shaping losses improve sample efficiency; the UNREAL framework [16] established that auxiliary predictive tasks help when extrinsic reward is sparse.

\subsection{Partial Observability and Belief Modeling}

In partially observable environments, optimal decision-making requires reasoning about the distribution over hidden states (the \textit{belief state}) [4]. Belief-state and information-set methods are powerful but heavyweight: POMCP [5] uses particle filtering for belief updates; IS-MCTS [6] operates over information sets; ReBeL [11] and Student of Games [12] combine learned models with game-theoretic search over public belief states; BetaZero [19] runs MuZero-style iteration directly in belief space. These approaches substantially improve performance under hidden information (e.g., poker [20, 21, 22]) but add significant algorithmic and computational cost. A lighter alternative is to treat MuZero's recurrent hidden state as an implicit belief representation. Our work fills the gap between full belief-state planning and unmodified MuZero: we add auxiliary outcome-prediction heads that \textit{shape} the latent state for partial observability without maintaining explicit belief distributions or changing the search algorithm.

\section{Environment Design}

\subsection{Skyjo Game Rules}

Skyjo is a card game for 2--8 players. The deck contains 150 cards with values ranging from $-2$ to $12$, distributed as follows: five cards of value $-2$, ten cards each of value $-1$ and values $1$ through $12$, and fifteen cards of value $0$.

\begin{figure}[ht]
\centering
\includegraphics[width=0.6\textwidth]{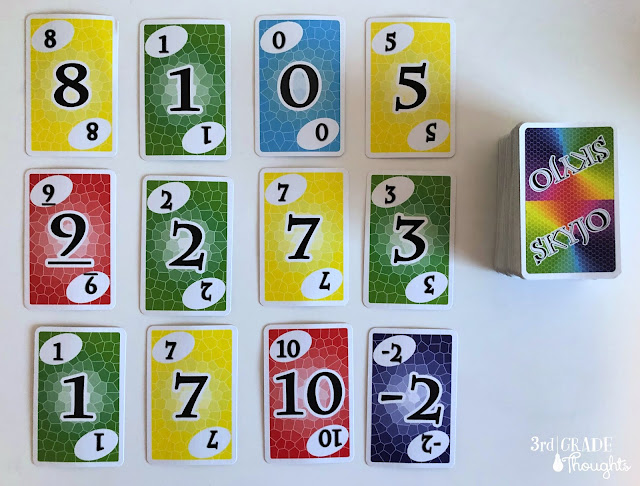}
\caption{A player's $3 \times 4$ card grid during a game of Skyjo, with the deck shown at right. Cards range from $-2$ to $12$, with lower total scores being more desirable. Image credit: 3rd Grade Thoughts (\url{https://www.3rdgradethoughts.com/2019/01/board-game-review-skyjo.html}).}
\label{fig:skyjo_board}
\end{figure}

Each player receives 12 cards arranged in a $3 \times 4$ grid, all initially face-down (see Figure~\ref{fig:skyjo_board} for an example of a player's grid during play). At the start of a round, each player reveals two cards. The player with the highest sum of revealed cards takes the first turn. On each turn, a player must either:

\begin{itemize}[nosep]
    \item Draw the top card from the discard pile and replace any card in their grid (the replaced card is discarded face-up), or
    \item Draw from the deck, observe the card's value, then either:
    \begin{itemize}[nosep]
        \item Keep the drawn card and replace any grid card, or
        \item Discard the drawn card and flip one face-down card face-up.
    \end{itemize}
\end{itemize}

When all three cards in a column are face-up and share the same value, that column is removed from the grid and no longer contributes to the player's score. When a player has all remaining cards face-up, the round ends: each other player gets one final turn. The round-ending player's score is doubled if they do not hold the strictly lowest score. Play continues across rounds until any player's cumulative score reaches 100; the player with the lowest cumulative score wins.

\subsection{Decision-Granularity Action Decomposition}

A naive approach to modeling Skyjo turns would compress an entire turn into a single macro-action. However, this is fundamentally flawed because the value of a card drawn from the deck is unknown at the time the draw decision is made, so the agent must \textit{observe} the drawn card before deciding whether to keep or discard it. This is the same insight that motivates Stochastic MuZero's [7] separation of decision nodes from chance nodes, and Sampled MuZero's [17] decoupling of available actions from actions considered in search.

We decompose each turn into a sequence of decision points:

\begin{itemize}
    \item \textbf{Phase A -- Choose Source:} The agent selects between the deck and the discard pile ($|\mathcal{A}| = 2$).
    \item \textbf{Phase B -- Keep or Discard:} If the agent drew from the deck, the card value is revealed. The agent then decides to keep or discard the drawn card ($|\mathcal{A}| = 2$).
    \item \textbf{Phase C -- Choose Position:} The agent selects a grid position for replacement or flipping ($|\mathcal{A}| = 12$).
\end{itemize}

This yields a masked 16-action policy space where legal actions depend on the current decision phase. The decomposition ensures correct information flow: the agent's keep-or-discard decision is conditioned on the observed card value, and MCTS can branch meaningfully at each decision point.

\subsection{Observation Model}

The environment provides each agent with a partial observation consisting of:

\begin{itemize}[nosep]
    \item \textbf{Board tokens:} For each player, 12 tokens encoding position, owner, visibility flag, and card value (or a special \texttt{UNKNOWN} sentinel for face-down cards).
    \item \textbf{Discard token:} The top card value and discard pile size.
    \item \textbf{Global token:} Deck size, step count, current player, turn phase, round phase bucket (early/mid/late), and round index.
    \item \textbf{Action history tokens:} The last $K = 16$ public actions, each encoding the actor, action type, source, target position, and card values involved.
    \item \textbf{Decision token:} The current decision phase and the drawn card value (if applicable).
\end{itemize}

This token-based representation is designed for direct consumption by a transformer encoder and scales naturally with the number of players.

\section{Methodology}

\subsection{Classical MuZero Baseline}

\begin{figure}[ht]
\centering
\includegraphics[width=\textwidth]{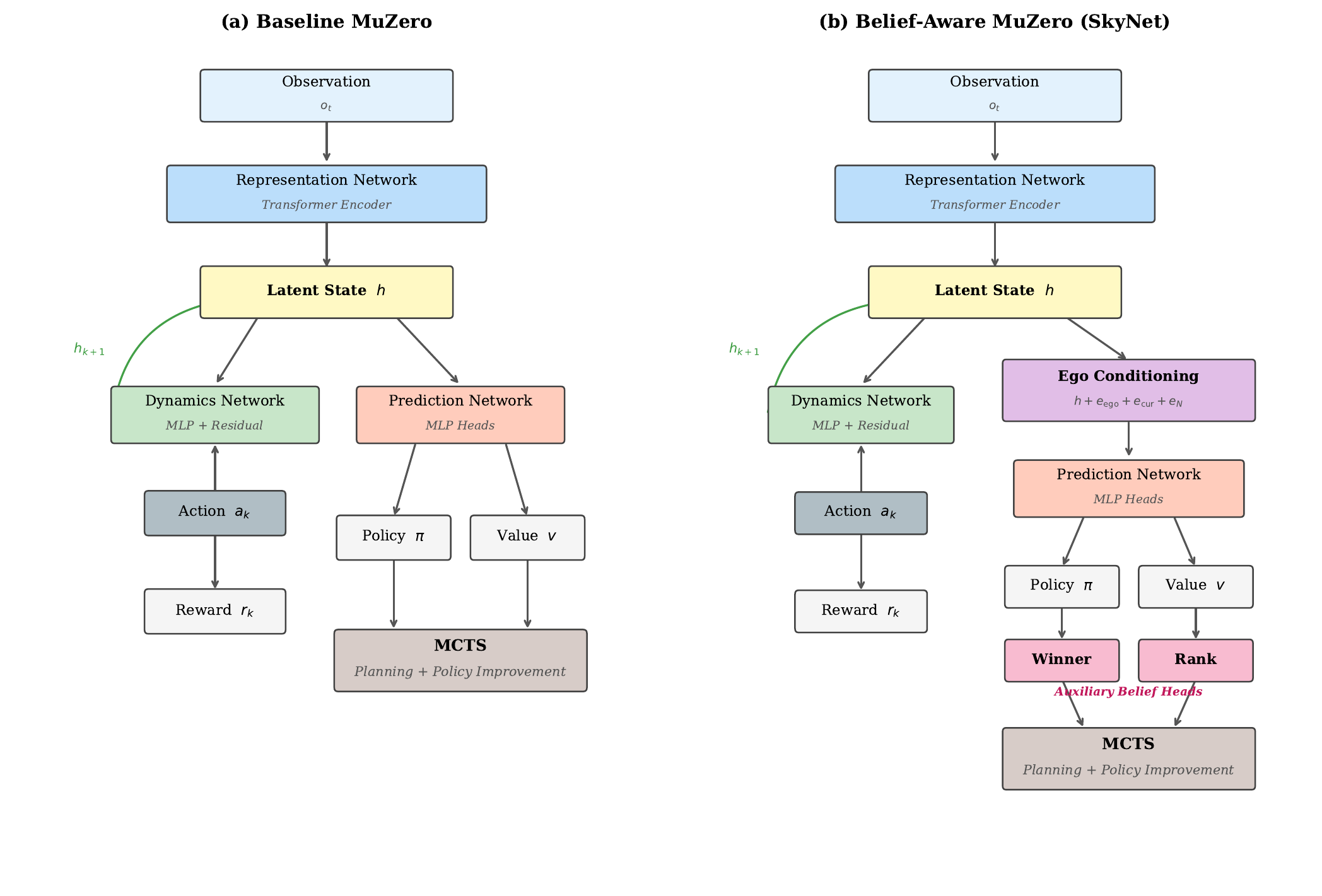}
\caption{Architecture comparison between baseline MuZero (left) and Belief-Aware MuZero / SkyNet (right). Both share the same representation, dynamics, and prediction networks. SkyNet adds an ego conditioning layer that injects player identity before prediction, and two auxiliary heads (winner and rank) that shape the latent representation via outcome-prediction objectives. Pink-shaded outputs indicate the additional belief heads.}
\label{fig:architecture}
\end{figure}

Our baseline follows the standard MuZero architecture [1], which learns a latent dynamics model of the environment without access to its rules. Unlike model-free approaches that map observations directly to actions, MuZero constructs an internal model that can be used for lookahead planning via MCTS. The key insight, formalized by the value-equivalence principle [14], is that this internal model does not need to reconstruct observations faithfully; it only needs to support accurate value and policy predictions. The architecture decomposes into three learned components that together define a self-contained planning engine in latent space. Figure~\ref{fig:architecture} illustrates the baseline architecture alongside the belief-aware extension described in Section~4.2.

\subsubsection{Representation Network}

The representation network $h_0 = f_{\text{rep}}(o_t)$ maps a tokenized observation to a latent state vector. We use a transformer encoder operating over the concatenation of board tokens, discard tokens, global tokens, action history tokens, and decision tokens.

Each token type is embedded through dedicated embedding layers. Board tokens combine embeddings for owner, position, visibility, and value. A learnable \texttt{[CLS]} token is prepended, and positional embeddings are added to the full sequence. The transformer encoder consists of 6 layers with 8 attention heads, GELU activations, and pre-norm architecture. The \texttt{[CLS]} output is projected through a layer norm and linear layer with Tanh activation to produce a latent vector $h_0 \in \mathbb{R}^{512}$.

\subsubsection{Dynamics Network}

The dynamics network $h_{k+1}, r_k = f_{\text{dyn}}(h_k, a_k)$ predicts the next latent state and reward given the current latent state and an action. The action is embedded via a learned embedding, concatenated with the current latent state, and processed through a two-hidden-layer MLP with GELU activations. A residual connection adds the MLP output to the input latent state, followed by layer normalization. A separate reward head predicts the reward as a categorical distribution over a discrete support.

\subsubsection{Prediction Network}

The prediction network $\pi_k, v_k = f_{\text{pred}}(h_k)$ produces policy logits over the action space and a value estimate from the latent state. Both outputs are produced by separate MLPs. The value is represented as a categorical distribution over the support $[-V_{\max}, V_{\max}]$ with $V_{\max} = 200$. In this non-zero-sum multiplayer setting, the value head predicts the expected discounted return from the \textit{ego} player's perspective (the player whose turn it is at the root of the current search). Training uses the same terminal reward (win/loss or score-based outcome) for that player; the value is not a zero-sum payoff and does not invert between players. MCTS uses this ego-conditioned value only for the acting player when comparing actions.

\subsubsection{Training}

Training follows the standard MuZero procedure [1]: MCTS is used during self-play to generate improved policy targets from visit count distributions, and $n$-step bootstrapped returns [18] provide value targets. The loss combines cross-entropy for policy and value heads and reward prediction:

\begin{equation}
\mathcal{L} = \sum_{k=0}^{K} \left( \mathcal{L}_{\pi}^{(k)} + \mathcal{L}_{v}^{(k)} \right) + \sum_{k=0}^{K-1} \mathcal{L}_{r}^{(k)} + \lambda \|\theta\|^2
\end{equation}

where $K$ is the unroll length and $\lambda$ controls weight decay regularization.

\subsection{Belief-Aware MuZero}

The Belief-Aware MuZero variant extends the baseline with two key modifications: auxiliary prediction heads for winner and rank estimation, and ego-conditioned latent representations. This approach draws on the principle that auxiliary tasks improve representations when extrinsic reward is sparse [10, 16], while the ego conditioning mechanism is inspired by multi-player search methods that maintain player-specific value estimates [11, 12].

\subsubsection{Ego Conditioning}

In multi-player partially observable games, the value of a state depends critically on \textit{whose perspective} it is evaluated from. We introduce ego conditioning by adding three learned embeddings to the latent state before prediction:

\begin{equation}
h_{\text{cond}} = \text{LayerNorm}\left(h + e_{\text{ego}} + e_{\text{current}} + e_{\text{nplayers}}\right)
\end{equation}

where $e_{\text{ego}}$ is the embedding of the ego player (the player whose value we are estimating), $e_{\text{current}}$ is the embedding of the player currently acting, and $e_{\text{nplayers}}$ encodes the total number of players. This conditioning allows a single network to produce player-specific predictions, enabling perspective augmentation during training where each trajectory generates training examples from every player's viewpoint.

\needspace{22\baselineskip}
\subsubsection{Winner and Rank Heads}

Two auxiliary prediction heads are added:

\begin{itemize}[nosep]
    \item \textbf{Winner head:} Predicts $P(\text{player } i \text{ finishes first})$ for each player, trained with cross-entropy loss against the observed game outcome.
    \item \textbf{Rank head:} Predicts the expected final rank distribution for each player, providing a denser training signal than the sparse binary win indicator.
\end{itemize}

Both heads operate on the ego-conditioned latent state and share the same MLP architecture as the policy and value heads. The total loss becomes:

\begin{equation}
\mathcal{L}_{\text{belief}} = \mathcal{L}_{\text{MuZero}} + \alpha \cdot \mathcal{L}_{\text{winner}} + \beta \cdot \mathcal{L}_{\text{rank}}
\end{equation}

where $\alpha$ and $\beta$ are loss weights that follow a ramping schedule to prevent auxiliary tasks from dominating early training.

\subsubsection{Rationale}

The auxiliary heads serve two purposes. First, they encourage the latent state to retain information useful for outcome prediction under partial observability, plausibly including hidden-card and opponent-state structure. This acts as a form of self-supervised regularization on the latent space, analogous to the representation-shaping objectives in EfficientZero [10] and SPR [15]. Second, the winner head provides a direct estimate of win probability that can be used alongside the shaped value estimate during planning.

\subsection{Training Pipeline}

\subsubsection{Self-Play with MCTS}

Both models use MCTS [3] during self-play to generate training data. At each decision point, the agent runs $N_{\text{sim}}$ simulations using the PUCT selection criterion with Dirichlet noise at the root for exploration. The visit count distribution provides the policy target, and the search value provides bootstrapped value targets.

For the belief-aware model, MCTS incorporates ego conditioning: the ego player ID is propagated through dynamics unrolls so that value estimates remain perspective-consistent.

\subsubsection{Curriculum and Opponent Pool}

Training employs a curriculum of six hand-crafted heuristic opponents: \textit{greedy value replacement}, \textit{information-first flip}, \textit{column hunter}, \textit{risk-aware unknown replacement}, \textit{end-round aggro}, and \textit{anti-discard}. These bots exploit different strategic principles of Skyjo and provide diverse training signal during early iterations.

As training progresses, a checkpoint opponent pool accumulates past versions of the agent. Self-play opponents are sampled from a mixture of the current policy (70\%) and uniformly from the pool (30\%), preventing the self-play collapse and oscillation that arises from training exclusively against the latest policy.

\subsubsection{Replay Buffer and Sampling}

Episodes are stored in a large replay buffer with phase-stratified sampling to ensure representation of all decision phases. Prioritized experience replay [23] principles guide sampling, with the buffer capacity set to accommodate thousands of episodes and a minimum warmup threshold ensuring sufficient data diversity before training begins.

\subsubsection{Simulation Schedule}

MCTS simulation counts follow a schedule that increases over training: 200 simulations per move for iterations 0--200, 400 for iterations 200--500, and 600 for iterations 500+. This balances computational cost against target quality, following the insight from Gumbel MuZero [8] that policy improvement quality depends critically on simulation budget.

\subsubsection{Time Penalty}

The training pipeline supports an optional per-step penalty $\lambda_t$ added to rewards to discourage pathologically long games. In early experiments with this penalty enabled, both models would frequently collapse to policies that optimized only for the fastest possible game termination, a degenerate strategy that did not restabilize with continued training. The current models were trained without this penalty ($\lambda_t = 0$); the opponent pool and curriculum diversity proved sufficient to maintain healthy game lengths without requiring explicit time pressure.

\section{Experimental Setup}

\subsection{Hyperparameters}

Table~\ref{tab:hyperparams} summarizes the key hyperparameters shared between models and those specific to each variant.

\begin{table}[H]
\centering
\caption{Training hyperparameters.}
\label{tab:hyperparams}
\small
\begin{tabular}{lcc}
\toprule
\textbf{Parameter} & \textbf{Baseline} & \textbf{Belief-Aware} \\
\midrule
Iterations & 2000 & 2000 \\
Self-play episodes/iter & 32 & 32 \\
Train steps/iter & 96 & 96 \\
Batch size & 64 & 64 \\
Unroll steps / TD steps & 8 / 10 & 8 / 10 \\
Discount ($\gamma$) & 0.997 & 0.997 \\
Learning rate & $3 \times 10^{-4}$ & $3 \times 10^{-4}$ \\
Optimizer & AdamW & AdamW \\
\midrule
Transformer layers / heads & 6 / 8 & 6 / 8 \\
$d_{\text{model}}$ / Latent dim & 256 / 512 & 256 / 512 \\
FF hidden dim & 1024 & 1024 \\
\midrule
MCTS sims (self-play / eval) & 200--600 / 200 & 200--600 / 200 \\
Dirichlet $\alpha$ / fraction & 0.3 / 0.25 & 0.3 / 0.25 \\
\midrule
Winner loss weight ($\alpha$) & --- & $0.1 \to 0.5$ \\
Rank loss weight ($\beta$) & --- & $0.1 \to 0.25$ \\
Action space & 16 & 16 \\
\bottomrule
\end{tabular}
\end{table}

\needspace{14\baselineskip}
\subsection{Evaluation Protocol}

Models are evaluated through two mechanisms:

\begin{enumerate}
    \item \textbf{Bot evaluation:} Periodic evaluation against the curriculum of heuristic opponents, measuring win rate, mean score differential, and mean episode length.
    \item \textbf{Head-to-head comparison:} Direct competition between baseline and belief-aware checkpoints at matched training iterations, with alternating seat assignments and 1000 games per comparison.
\end{enumerate}

Win rate is determined by final Skyjo score (lowest score wins), computed regardless of whether the episode terminated naturally or was truncated. Truncation rate is tracked separately to ensure evaluation integrity. Evaluation uses greedy MCTS (temperature $\approx 0$) with 200 simulations per move and no exploration noise. This protocol follows best practices from game evaluation frameworks such as OpenSpiel [24] and RLCard [25].

\subsection{Comparison Fairness}

The baseline and belief-aware models are compared under matched conditions. Both use the same training pipeline (self-play episodes per iteration, replay buffer, curriculum, opponent pool), the same core hyperparameters (learning rate, batch size, unroll steps, discount, MCTS simulation schedule), and the same evaluation protocol (200 sims per move, alternating seats, 1000 games per head-to-head). No additional hyperparameter tuning was performed for the belief-aware variant beyond the auxiliary loss weight ramp. Checkpoint selection uses the same iteration-matching protocol: head-to-head comparisons use the same training iteration for both agents. The belief-aware model adds only the ego conditioning embeddings and two auxiliary MLP heads (winner and rank) to the shared representation and dynamics; parameter counts are comparable. Thus the performance difference is attributable to the architectural change rather than capacity or tuning advantage.

\section{Usage of ROSIE}

Training MuZero-style agents on a complex partially observable game is computationally demanding: each iteration involves running MCTS-guided self-play episodes, updating the replay buffer, and performing gradient steps through a transformer-based representation network and unrolled dynamics. At this scale, compute access is a hard constraint on experimental velocity, not a convenience.

Training on MSOE's T4 teaching-cluster nodes required approximately 24 hours per 100 training iterations. Running the full 2000-iteration curriculum on such hardware alone would take roughly 480 hours (20 days) \textit{per model}, serial, before any ablations or head-to-head evaluations could be conducted. MSOE's ROSIE supercomputer provides access to DGX H100 nodes, which achieved approximately 1.25$\times$ the throughput of the T4 nodes, reducing wall-clock time meaningfully. Attempting to train locally, whether on CPU or a consumer-grade GPU, produced iteration times at roughly 0.88$\times$ the speed of the teaching nodes, making sustained multi-thousand-iteration runs infeasible in practice.

Beyond raw throughput, ROSIE enabled a qualitatively different development workflow. Because every additional iteration translates directly into additional development cycles and the ability to test new hypotheses, the ability to run training at scale compressed the experimental feedback loop substantially. Critically, ROSIE made it possible to \textit{parallelize} the research program: the baseline MuZero and SkyNet models were trained simultaneously on separate nodes, head-to-head evaluations between checkpoints were run concurrently with ongoing training, and inference-time ablation experiments were conducted in parallel without interrupting either training job. This parallelism was essential for completing the matched-checkpoint head-to-head comparisons (Table~\ref{tab:h2h}) and the inference-time ablation study (Table~\ref{tab:ablation}) within a practical project timeline.

\section{Results}

\subsection{Training Dynamics}

Both models exhibit healthy training dynamics: total loss decreases monotonically, gradient norms stabilize in the range of 1--3 after initial transients, and policy entropy decreases gradually without premature collapse.

The belief-aware model shows additional loss components that converge meaningfully. The winner head loss decreases from approximately 1.16 to below 0.02 over training, indicating that the network learns to predict game outcomes with increasing accuracy. The rank head loss follows a similar trajectory.

\begin{figure}[ht]
\centering
\includegraphics[width=\textwidth]{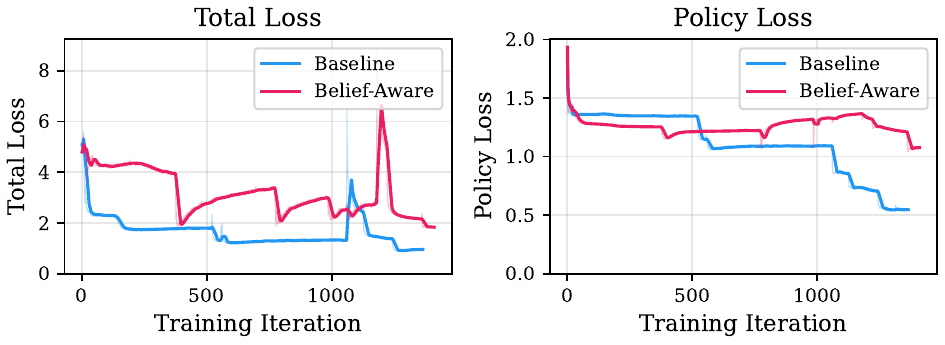}
\caption{Training loss curves for baseline MuZero and Belief-Aware MuZero. Raw values shown in light color; 20-iteration rolling averages in bold. Both models converge, with the belief-aware model's total loss higher due to the additional auxiliary loss terms.}
\label{fig:training_loss}
\end{figure}

\begin{figure}[ht]
\centering
\includegraphics[width=\textwidth]{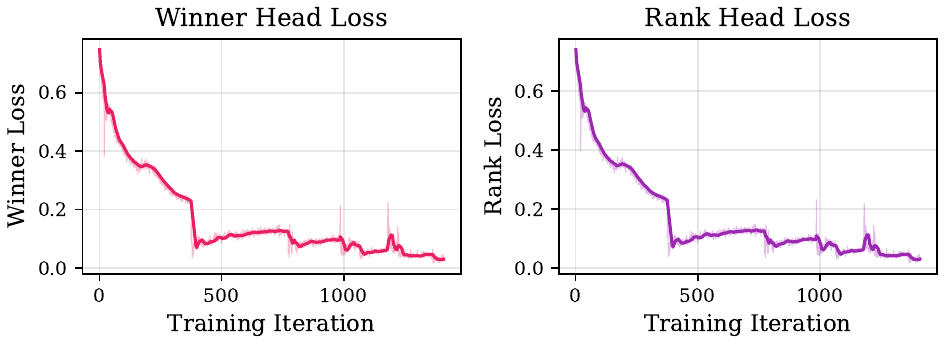}
\caption{Convergence of the belief-aware model's auxiliary prediction heads. The winner head loss (left) and rank head loss (right) both decrease substantially over training, confirming that the network learns to predict game outcomes with increasing accuracy.}
\label{fig:auxiliary_loss}
\end{figure}

\subsection{Self-Play Evaluation}

During training with the scaled pipeline (32 self-play episodes per iteration, opponent pool, simulation schedule), the belief-aware model consistently achieves higher evaluation win rates against the heuristic bot curriculum. Table~\ref{tab:eval_comparison} summarizes evaluation statistics computed over all evaluation checkpoints.

\begin{table}[H]
\centering
\caption{Evaluation statistics over training (self-play vs.\ heuristic opponents). Bracketed ranges: $\mu \pm 1.96\sigma$ across checkpoints.}
\label{tab:eval_comparison}
\begin{tabular}{lcc}
\toprule
\textbf{Metric} & \textbf{Baseline} & \textbf{Belief-Aware} \\
\midrule
Mean eval win rate ($\pm 1.96\sigma$) & 0.466 [0.32--0.61] & 0.720 [0.58--0.87] \\
Std eval win rate & 0.073 & 0.074 \\
Max eval win rate & 0.600 & 0.825 \\
Min eval win rate & 0.313 & 0.525 \\
Mean truncation rate & 0.002 & 0.000 \\
Mean episode length & 87.5 & 87.7 \\
\bottomrule
\end{tabular}
\end{table}

The belief-aware model achieves a mean win rate of 0.720 compared to the baseline's 0.466. The near-identical episode lengths and negligible truncation rates confirm that this performance difference is not attributable to tempo manipulation or evaluation artifacts. The belief-aware model's minimum observed win rate exceeds the baseline's mean across checkpoints, further supporting the consistency of the improvement. These results corroborate the head-to-head findings reported in Section~\ref{sec:h2h}.

\begin{figure}[H]
\centering
\includegraphics[width=0.8\textwidth]{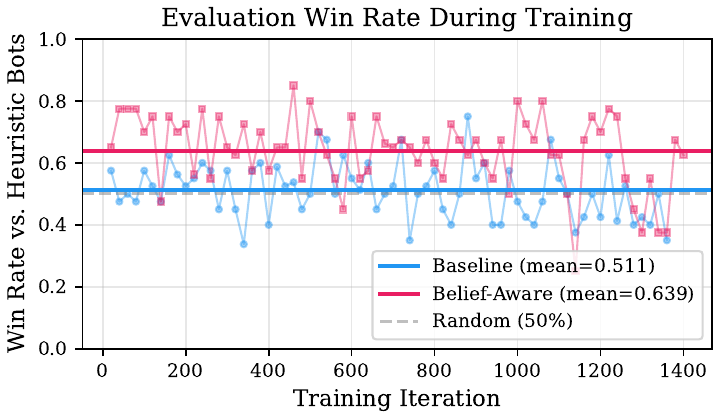}
\caption{Evaluation win rate against the heuristic bot curriculum over training. The belief-aware model consistently achieves higher win rates after an initial ramp-up period. The dashed line indicates the 50\% random baseline.}
\label{fig:eval_winrate}
\end{figure}

\subsection{Head-to-Head Results}
\label{sec:h2h}

Table~\ref{tab:h2h} presents head-to-head results between the two architectures at various matched checkpoints, each evaluated over 1000 games.

\begin{table}[H]
\centering
\caption{Head-to-head results (1000 games per comparison). Belief WR 95\% CI: Wilson score interval.}
\label{tab:h2h}
\begin{tabular}{lccccc}
\toprule
\textbf{Checkpoint} & \textbf{Belief Wins} & \textbf{Baseline Wins} & \textbf{Draws} & \textbf{Belief WR (95\% CI)} & \textbf{$\Delta$ Elo} \\
\midrule
Iter 125 & 360 & 640 & 0 & 36.0\% [33.1--39.0] & $-99$ \\
Iter 250 & 422 & 578 & 0 & 42.2\% [39.2--45.3] & $-55$ \\
Iter 500 & 668 & 332 & 0 & 66.8\% [63.8--69.6] & $+120$ \\
Iter 750 & 742 & 252 & 6 & 74.2\% [71.4--76.8] & $+184$ \\
Iter 1000 & 753 & 240 & 7 & 75.3\% [72.5--77.9] & $+194$ \\
\bottomrule
\end{tabular}
\end{table}

\begin{figure}[H]
\centering
\includegraphics[width=0.8\textwidth]{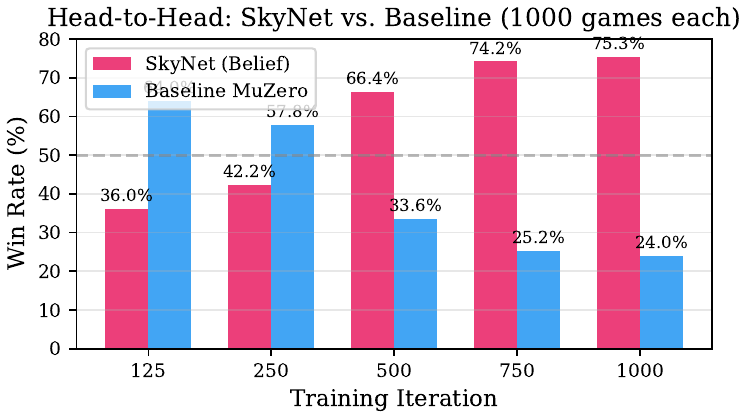}
\caption{Head-to-head win rates between SkyNet (Belief-Aware) and baseline MuZero at matched training checkpoints (1000 games each). The crossover between iterations 250 and 500 shows SkyNet overtaking the baseline after sufficient training.}
\label{fig:h2h}
\end{figure}

A clear crossover occurs between iterations 250 and 500: the belief-aware model initially underperforms the baseline (likely due to the additional parameter complexity and auxiliary loss overhead during early training), but surpasses it decisively once sufficient training data has been accumulated. The advantage peaks at iteration 1000 with a 75.3\% win rate, corresponding to +194 Elo [26]. Figure~\ref{fig:h2h} visualizes this trajectory.

\subsection{Statistical Significance}

For the peak at iteration 1000 (753 wins out of 1000 games), under the null hypothesis that both models are equally strong ($p = 0.5$):
\begin{equation}
z = \frac{0.753 - 0.5}{\sqrt{0.5 \cdot 0.5 / 1000}} = \frac{0.253}{0.0158} \approx 16.0
\end{equation}

This yields $p < 10^{-50}$, providing overwhelming statistical evidence that the belief-aware model is stronger at this checkpoint. The 95\% Wilson score interval for the iter 1000 win rate is [72.5\%, 77.9\%], excluding 50\%.

At iteration 500, the win rate of 66.8\% (668/1000) yields $z \approx 10.6$ ($p < 10^{-25}$), with 95\% Wilson interval [63.8\%, 69.6\%], confirming that the advantage is already decisive before reaching the peak.

\subsection{Inference-Time Ablation}

Because full retraining ablations were not feasible within the project timeline, we use an inference-time ablation to isolate whether the trained policy actively relies on ego conditioning during planning. The same trained SkyNet checkpoint (iter 500) is evaluated in two modes: (i)~full inference with ego conditioning and (ii)~ablated inference with ego conditioning disabled. The model weights are identical; only the prediction path differs (bypassing the ego, current-player, and number-of-players embeddings).

\begin{table}[H]
\centering
\caption{Inference-time ablation: full SkyNet vs.\ ego-conditioning-off (1000 games, iter 500). WR 95\% CI: Wilson score interval.}
\label{tab:ablation}
\small
\begin{tabular}{lccccc}
\toprule
\textbf{Checkpoint} & \textbf{Full Wins} & \textbf{Ablated Wins} & \textbf{Draws} & \textbf{Full WR (95\% CI)} & \textbf{$\Delta$ Elo} \\
\midrule
Iter 500 & 690 & 305 & 5 & 69.0\% [66.1--71.8] & $+139$ \\
Iter 1000 & 811 & 181 & 8 & 81.1\% [78.6--83.4] & $+253$ \\
\bottomrule
\end{tabular}
\end{table}

At iteration 500, the full model wins 69.0\% of 1000 games against the ablated variant ($z = 12.0$, $p < 10^{-30}$). At iteration 1000, this effect strengthens: the full model wins 81.1\% of 1000 games ($z \approx 19.7$, $p < 10^{-50}$). These results suggest that ego-conditioned predictions contribute directly to decision quality during planning rather than serving only as a training-time representation shaping signal. Combined with the baseline comparison (Section~\ref{sec:h2h}), this shows that the performance gain is not solely attributable to auxiliary-loss regularization; the model actively exploits ego conditioning at inference time.

This is an inference-time ablation only and does not separate the individual contributions of training-time auxiliary supervision from ego conditioning during training. A full training ablation (ego-only, heads-only) remains future work.

\subsection{Training Stability}

Early experiments with limited self-play throughput (4 episodes per iteration) revealed severe instability in both models, with win rates oscillating between 0.25 and 0.875 across evaluation points. The belief-aware model exhibited higher variance than the baseline under these conditions, suggesting that the additional model complexity amplifies sensitivity to data scarcity.

\begin{figure}[H]
\centering
\includegraphics[width=\textwidth]{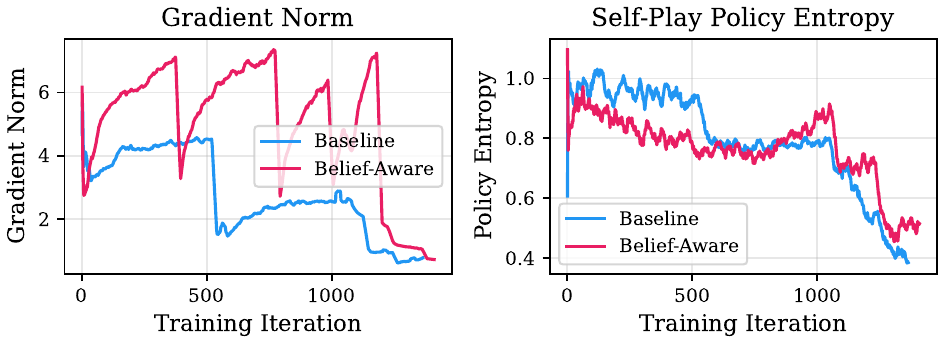}
\caption{Training diagnostics for both models (20-iteration rolling averages). Gradient norms (left) stabilize in a healthy range for both models. Policy entropy (right) decreases gradually without premature collapse, indicating progressive strategy refinement.}
\label{fig:diagnostics}
\end{figure}

Scaling self-play throughput to 32 episodes per iteration, introducing the opponent pool, and applying the simulation schedule eliminated this instability. Under the scaled pipeline, both models' evaluation win rate standard deviations are approximately 0.074, indicating that the belief-aware model's advantage is not achieved at the cost of stability. Figure~\ref{fig:diagnostics} shows that gradient norms and policy entropy evolve similarly for both models, confirming stable training dynamics.

\subsection{Latent Representation Analysis}

\begin{figure}[H]
\centering
\includegraphics[width=\textwidth]{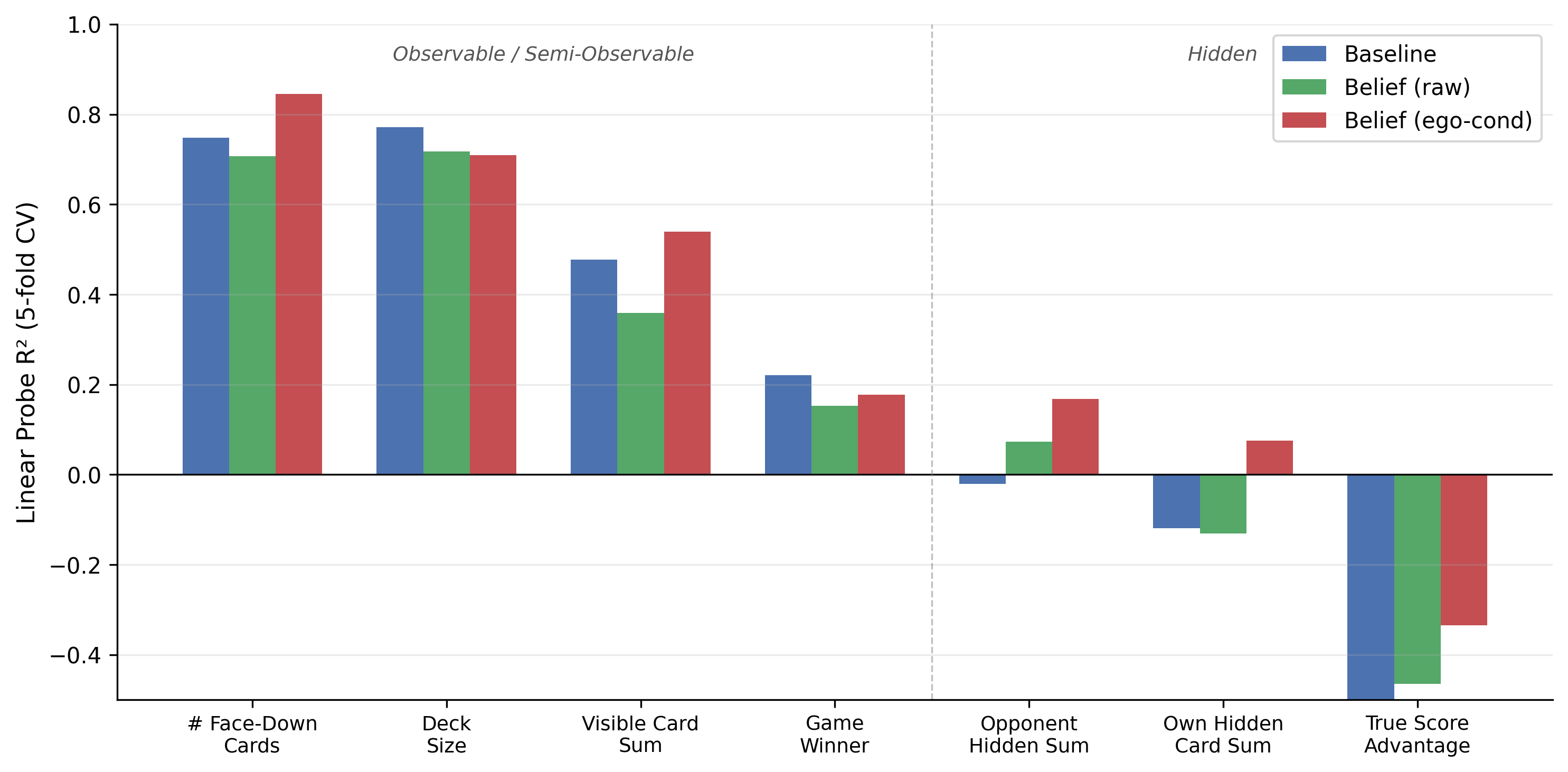}
\caption{Linear probe $R^2$ on frozen latent vectors for seven game-state features. Features left of the dashed line are directly observable or semi-observable; features to the right are hidden. Negative $R^2$ indicates worse-than-mean prediction. The ego-conditioned belief representation is the only variant that achieves positive $R^2$ on both hidden-card features.}
\label{fig:linear_probe}
\end{figure}

To understand \textit{how} the auxiliary heads shape the latent space, we train linear probes (ridge regression, 5-fold cross-validation) on frozen latent vectors from 44{,}179 decision steps across 150 games. We probe three representations: the baseline latent state, the belief-aware latent state before ego conditioning, and the belief-aware latent state after ego conditioning. Figure~\ref{fig:linear_probe} summarizes how much hidden-state information is linearly decodable from each.

All three representations encode observable features well: the number of face-down cards ($R^2 = 0.75$--$0.85$), deck size ($R^2 \approx 0.71$--$0.77$), and visible card sum ($R^2 = 0.36$--$0.54$) are linearly recoverable. The ego-conditioned representation achieves the highest $R^2$ on face-down card count ($0.845$ vs.\ baseline $0.748$) and visible sum ($0.539$ vs.\ $0.477$), indicating that ego conditioning sharpens the encoding of player-specific observable state.

The more striking result involves genuinely hidden features. Neither the baseline nor the raw belief representation achieves positive $R^2$ on the sum of face-down card values or opponent hidden card sum. The ego-conditioned representation, however, is the only variant that crosses into positive territory on both: $R^2 = 0.076$ for own hidden card sum and $R^2 = 0.168$ for opponent hidden sum. While these values are modest, they demonstrate that the auxiliary heads encourage the latent state to retain linear traces of hidden information that the baseline discards entirely. Across all hidden features, the belief models produce consistently less negative $R^2$ than the baseline (e.g., true score advantage: $-0.334$ vs.\ $-1.062$), suggesting improved alignment with hidden-state structure even where full linear decodability is not achieved.

These findings support the interpretation advanced in Section~7.1 that the auxiliary heads act as representation-shaping regularizers, encouraging the latent space to retain outcome-relevant hidden information without requiring explicit belief-state reconstruction.

\section{Discussion}

\subsection{Why Belief Modeling Helps}

The belief-aware model's advantage likely stems from two mechanisms. First, the auxiliary winner and rank heads impose an inductive bias on the shared representation: to predict game outcomes accurately, the latent state is encouraged to retain information relevant under partial observability, plausibly including hidden-card structure, deck composition, and opponent state. This acts as a form of self-supervised regularization that improves the quality of latent states used for both value estimation and MCTS planning. This is consistent with findings from EfficientZero [10] and the value-equivalence principle [14] that representations should be optimized for planning-relevant quantities rather than observation reconstruction.

Second, the ego conditioning mechanism allows the network to produce player-specific predictions from a shared representation, enabling effective perspective augmentation during training. Each self-play trajectory generates training signal from every player's viewpoint, effectively multiplying the data efficiency by the number of players.

\needspace{14\baselineskip}
\subsection{The Crossover Effect}

The observation that the belief-aware model \textit{underperforms} the baseline at early checkpoints (iterations 125--250) but \textit{dominates} later (iterations 500+) is consistent with a well-known pattern in multi-task and auxiliary-loss learning: the additional loss terms compete with the primary MuZero objective for gradient bandwidth during early training when data is scarce. The ramping loss weight schedule ($\alpha: 0.1 \to 0.5$, $\beta: 0.1 \to 0.25$) mitigates but does not eliminate this effect.

This finding has practical implications: practitioners should not evaluate belief-augmented architectures too early in training, as initial underperformance may be followed by substantial gains.

\subsection{Non-Zero-Sum Dynamics}

Skyjo is fundamentally non-zero-sum: players' scores are independent quantities, and one player's improvement does not necessarily correspond to another's decline. This distinguishes it from zero-sum games like Chess and Go where MuZero was originally validated. In non-zero-sum settings, self-play training is more prone to instability because value estimates cannot be cleanly inverted between players. The belief-aware model's explicit multi-player heads (winner prediction per player, rank estimation per player) provide a more natural interface for non-zero-sum value estimation than a single scalar value head, analogous to how ReBeL [11] and Student of Games [12] handle imperfect-information structure through explicit game-theoretic reasoning.

\subsection{The Role of Training Scale}

Perhaps the most important finding is that the belief-aware model's advantage only materializes under sufficient training throughput. With 4 self-play episodes per iteration, the additional model complexity produced more instability than benefit. With 32 episodes per iteration and an opponent pool, the same architecture produced consistent, statistically significant gains. This suggests that belief-aware auxiliary supervision requires adequate data flow to realize its potential, a practical constraint that mirrors findings in the broader model-based RL literature [27].

\vspace{-0.5\baselineskip}
\subsection{Limitations}

Several limitations should be noted. First, our belief heads predict game outcomes (winner, rank) rather than maintaining explicit beliefs over hidden state variables (e.g., distributions over face-down card values). Explicit belief-state planning, as in BetaZero [19] or POMCP [5], could further improve planning quality by enabling informed chance-node expansion in MCTS. Accordingly, we interpret SkyNet as a belief-aware representation shaping method, not an explicit belief-state planner. Second, our evaluation is limited to 2-player Skyjo; the model supports 2--8 players architecturally but scaled multi-player experiments remain future work. Third, comparison to human performance has not been conducted, though the models' ability to outperform hand-crafted heuristic opponents suggests competitive play quality.

\vspace{-\baselineskip}
\section{Future Work}
\vspace{-0.5\baselineskip}

Several directions could extend this work: training the belief head to predict explicit distributions over face-down card values and using these beliefs to weight chance branches in MCTS following Stochastic MuZero [7]; evaluating 3--8 player Skyjo, where win signal sparsity ($\sim 1/N$) and opponent modeling complexity increase; migrating to a fully asynchronous actor-learner pipeline [28] to further scale throughput; conducting controlled studies against human players; and applying the belief-aware framework to other imperfect-information games (e.g., Hanabi, Hearts) to assess generality.

\section{Conclusion}

\footnotetext{Skyjo Web Arena: \url{https://skyjo.artzima.dev/} \quad SkyNet Repository: \url{https://github.com/DevArtech/skynet}}

We presented Belief-Aware MuZero, an extension of the MuZero framework that augments the standard architecture with ego-conditioned auxiliary heads for winner prediction and rank estimation. Applied to Skyjo, a partially observable, stochastic, multi-player card game with non-zero-sum dynamics, the belief-aware variant achieves a peak 75.3\% win rate (+194 Elo) against the classical MuZero baseline in 1000-game head-to-head evaluation ($p < 10^{-50}$).

The key insight is that belief-aware auxiliary supervision appears to improve the quality of learned representations under partial observability, but this benefit requires sufficient training throughput to materialize. Under low-data regimes, the additional model complexity increases instability, whereas under adequate data flow it produces consistent and substantial gains.

These results suggest that MuZero can be effectively adapted to imperfect-information multi-player card games, and that even simple belief-aware auxiliary heads, applied without explicit hidden-state modeling, provide meaningful improvements. This opens a path toward stronger game-playing agents in the broad class of partially observable, stochastic, multi-player domains that more closely resemble real-world decision-making challenges.

\newpage
\section*{References}
\addcontentsline{toc}{section}{References}

\begin{enumerate}

\item Schrittwieser, J., Antonoglou, I., Hubert, T., Simonyan, K., Sifre, L., Schmitt, S., Guez, A., Lockhart, E., Hassabis, D., Graepel, T., Lillicrap, T., \& Silver, D. (2020). Mastering Atari, Go, Chess and Shogi by planning with a learned model. \textit{Nature}, 588(7839), 604--609.

\item Silver, D., Hubert, T., Schrittwieser, J., Antonoglou, I., Lai, M., Guez, A., Lanctot, M., Sifre, L., Kumaran, D., Graepel, T., Lillicrap, T., Simonyan, K., \& Hassabis, D. (2018). A general reinforcement learning algorithm that masters Chess, Shogi, and Go through self-play. \textit{Science}, 362(6419), 1140--1144.

\item Kocsis, L. \& Szepesv\'{a}ri, C. (2006). Bandit based Monte-Carlo planning. \textit{European Conference on Machine Learning (ECML)}, 282--293.

\item Kaelbling, L. P., Littman, M. L., \& Cassandra, A. R. (1998). Planning and acting in partially observable stochastic domains. \textit{Artificial Intelligence}, 101(1--2), 99--134.

\item Silver, D. \& Veness, J. (2010). Monte-Carlo planning in large POMDPs. \textit{Advances in Neural Information Processing Systems (NeurIPS)}, 23, 2164--2172.

\item Cowling, P. I., Powley, E. J., \& Whitehouse, D. (2012). Information set Monte Carlo tree search. \textit{IEEE Transactions on Computational Intelligence and AI in Games}, 4(2), 120--143.

\item Antonoglou, I., Schrittwieser, J., Ozair, S., Hubert, T., \& Silver, D. (2022). Planning in stochastic environments with a learned model. \textit{International Conference on Learning Representations (ICLR)}.

\item Danihelka, I., Guez, A., Schrittwieser, J., \& Silver, D. (2022). Policy improvement by planning with Gumbel. \textit{International Conference on Learning Representations (ICLR)}.

\item Schrittwieser, J., Hubert, T., Manber, A., Hassabis, D., \& Silver, D. (2021). Online and offline reinforcement learning by planning with a learned model. \textit{Advances in Neural Information Processing Systems (NeurIPS)}, 34.

\item Ye, W., Liu, S., Kurutach, T., Abbeel, P., \& Gao, Y. (2021). Mastering Atari games with limited data. \textit{Advances in Neural Information Processing Systems (NeurIPS)}, 34.

\item Brown, N., Bakhtin, A., Lerer, A., \& Gong, Q. (2020). Combining deep reinforcement learning and search for imperfect-information games. \textit{Advances in Neural Information Processing Systems (NeurIPS)}, 33.

\item Schmid, M., Moravc\'{i}k, M., Burch, N., Kadlec, R., Davidson, J., Waugh, K., Lisý, V., Bowling, M., Lanctot, M., \& Munos, R. (2023). Student of Games: A unified learning algorithm for both perfect and imperfect information games. \textit{Science Advances}, 9(46).

\item Whitehouse, D., Powley, E. J., \& Cowling, P. I. (2011). Determinization and information set Monte Carlo tree search for the card game Dou Di Zhu. \textit{IEEE Conference on Computational Intelligence and Games (CIG)}, 87--94.

\item Grimm, C., Barreto, A., Singh, S., \& Silver, D. (2020). The value equivalence principle for model-based reinforcement learning. \textit{Advances in Neural Information Processing Systems (NeurIPS)}, 33.

\item Schwarzer, M., Anand, A., Goel, R., Hjelm, R. D., Courville, A., \& Bachman, P. (2021). Data-efficient reinforcement learning with self-predictive representations. \textit{International Conference on Learning Representations (ICLR)}.

\item Jaderberg, M., Mnih, V., Czarnecki, W., Schaul, T., Leibo, J. Z., Silver, D., \& Kavukcuoglu, K. (2017). Reinforcement learning with unsupervised auxiliary tasks. \textit{International Conference on Learning Representations (ICLR)}.

\item Hubert, T., Schrittwieser, J., Antonoglou, I., Barekatain, M., Schmitt, S., \& Silver, D. (2021). Learning and planning in complex action spaces. \textit{International Conference on Machine Learning (ICML)}.

\item Sutton, R. S. \& Barto, A. G. (2018). \textit{Reinforcement Learning: An Introduction} (2nd ed.). MIT Press.

\item Moss, R. J., Corso, A., Caers, J., \& Kochenderfer, M. J. (2024). BetaZero: Belief-state planning for long-horizon POMDPs using learned approximations. \textit{arXiv preprint arXiv:2306.00249}.

\item Morav\v{c}\'{\i}k, M., Schmid, M., Burch, N., Lis\'{y}, V., Morrill, D., Bard, N., Davis, T., Waugh, K., Johanson, M., \& Bowling, M. (2017). DeepStack: Expert-level artificial intelligence in heads-up no-limit poker. \textit{Science}, 356(6337), 508--513.

\item Brown, N. \& Sandholm, T. (2017). Superhuman AI for heads-up no-limit poker: Libratus beats top professionals. \textit{Science}, 359(6374), 418--424.

\item Brown, N. \& Sandholm, T. (2019). Superhuman AI for multiplayer poker. \textit{Science}, 365(6456), 885--890.

\item Schaul, T., Quan, J., Antonoglou, I., \& Silver, D. (2016). Prioritized experience replay. \textit{International Conference on Learning Representations (ICLR)}.

\item Lanctot, M., Lockhart, E., Lespiau, J.-B., Zambaldi, V., Upadhyay, S., P\'{e}rolat, J., Srinivasan, S., Timbers, F., Tuyls, K., Omidshafiei, S., Hennes, D., Morrill, D., Muller, P., Eber, T., Duran-Martin, G., De Vylder, B., Munos, R., Abramson, J., Vinyals, O., \& Bowling, M. (2020). OpenSpiel: A framework for reinforcement learning in games. \textit{arXiv preprint arXiv:1908.09453}.

\item Zha, D., Lai, K.-H., Cao, Y., Huang, S., Wei, R., Guo, J., \& Hu, X. (2020). RLCard: A platform for reinforcement learning in card games. \textit{International Joint Conference on Artificial Intelligence (IJCAI)}.

\item Elo, A. E. (1978). \textit{The Rating of Chessplayers, Past and Present}. Arco Publishing.

\item Moerland, T. M., Broekens, J., Plaat, A., \& Jonker, C. M. (2023). Model-based reinforcement learning: A survey. \textit{Foundations and Trends in Machine Learning}, 16(1), 1--118.

\item Horgan, D., Quan, J., Budden, D., Barth-Maron, G., Hessel, M., van Hasselt, H., \& Silver, D. (2018). Distributed prioritized experience replay. \textit{International Conference on Learning Representations (ICLR)}.

\end{enumerate}

\end{document}